\documentclass[10pt,twocolumn,letterpaper]{article}

\usepackage{cvpr}              

\usepackage[dvipsnames]{xcolor}

\usepackage{multirow} 
\usepackage{amsmath} 
\usepackage{bbm} 
\usepackage{siunitx} 
\usepackage{kotex} 
\usepackage{array,multirow,graphicx}
\usepackage{float}
\usepackage{amsmath}
\usepackage{mathtools}
\usepackage{bbm}
\usepackage{hhline}
\usepackage{boldline}
\usepackage{tabularx}
\usepackage{float}
\usepackage{comment}
\usepackage{color}
\usepackage{amssymb}
\usepackage{booktabs}
\usepackage{multicol}
\usepackage{microtype}      
\usepackage{xcolor}         
\usepackage{algorithm}
\usepackage{algpseudocode}
\usepackage{pifont}
\usepackage{soul} 
\usepackage{subcaption}
\usepackage{wrapfig}
\usepackage{lipsum}
\usepackage{booktabs}
\usepackage{xcolor,colortbl}
\usepackage{nicematrix}
\usepackage{kotex}

\usepackage[pagebackref,breaklinks,colorlinks,citecolor=cvprblue]{hyperref}
\definecolor{cvprblue}{rgb}{0.21,0.49,0.74}

\def\paperID{11779} 
\def\confName{CVPR}
\def\confYear{2025}

\title{Temporal Alignment-Free Video Matching for Few-shot Action Recognition}

\author{
SuBeen Lee, WonJun Moon, Hyun Seok Seong, Jae-Pil Heo\thanks{Corresponding author}\\
Sungkyunkwan University\\
{\tt\small \{leesb7426, wjun0830, gustjrdl95, jaepilheo\}@skku.edu}
}

\begin{document}
\maketitle
\begin{abstract}
Few-Shot Action Recognition (FSAR) aims to train a model with only a few labeled video instances. A key challenge in FSAR is handling divergent narrative trajectories for precise video matching. While the frame- and tuple-level alignment approaches have been promising, their methods heavily rely on pre-defined and length-dependent alignment units (e.g., frames or tuples), which limits flexibility for actions of varying lengths and speeds. In this work, we introduce a novel TEmporal Alignment-free Matching (TEAM) approach, which eliminates the need for temporal units in action representation and brute-force alignment during matching. Specifically, TEAM represents each video with a fixed set of pattern tokens that capture globally discriminative clues within the video instance regardless of action length or speed, ensuring its flexibility. Furthermore, TEAM is inherently efficient, using token-wise comparisons to measure similarity between videos, unlike existing methods that rely on pairwise comparisons for temporal alignment. Additionally, we propose an adaptation process that identifies and removes common information across classes, establishing clear boundaries even between novel categories. Extensive experiments demonstrate the effectiveness of TEAM. Codes are available at \href{github.com/leesb7426/TEAM}{github.com/leesb7426/TEAM}.
\end{abstract}
\section{Introduction}
\label{sec:intro}
Recently, action recognition has achieved remarkable performances with the help of numerous labeled videos~\cite{VIVIT, Kinetics, SlowFast}.
However, those performances are not guaranteed when novel categories are given with few labeled examples~\cite{ARN, CMN, OTAM, ProtoNet}.
To resolve this problem, Few-Shot Action Recognition~(FSAR) has been proposed~\cite{CMN}. 
Particularly, FSAR employs episodic learning which iteratively simulates the episode of learning a set of new classes.
Each episode is given a support set to train a model, composed of a few labeled examples for newly given classes, to distinguish the category of a set of unlabeled videos, coined as a query set.

Due to different class compositions within every episode, metric learning, a classifier-free strategy, is predominantly employed in FSAR~\cite{OTAM, HyRSM, MoLo, TRX, STRM, SloshNet, GgHM}.
Distances between the representations of support and query sets are computed to infer the category of a query video.
Yet, na\"ively constructing video representations, such as average pooling, are shown to be poor due to an insufficient understanding of temporal information~\cite{CMN, OTAM}.
Therefore, the previous approaches typically focused on aligning temporal sequences between videos~\cite{OTAM, HyRSM, MoLo, TRX, STRM, SloshNet, GgHM}.
\begin{figure*}
    \centering
    {\includegraphics[width=1.\textwidth]{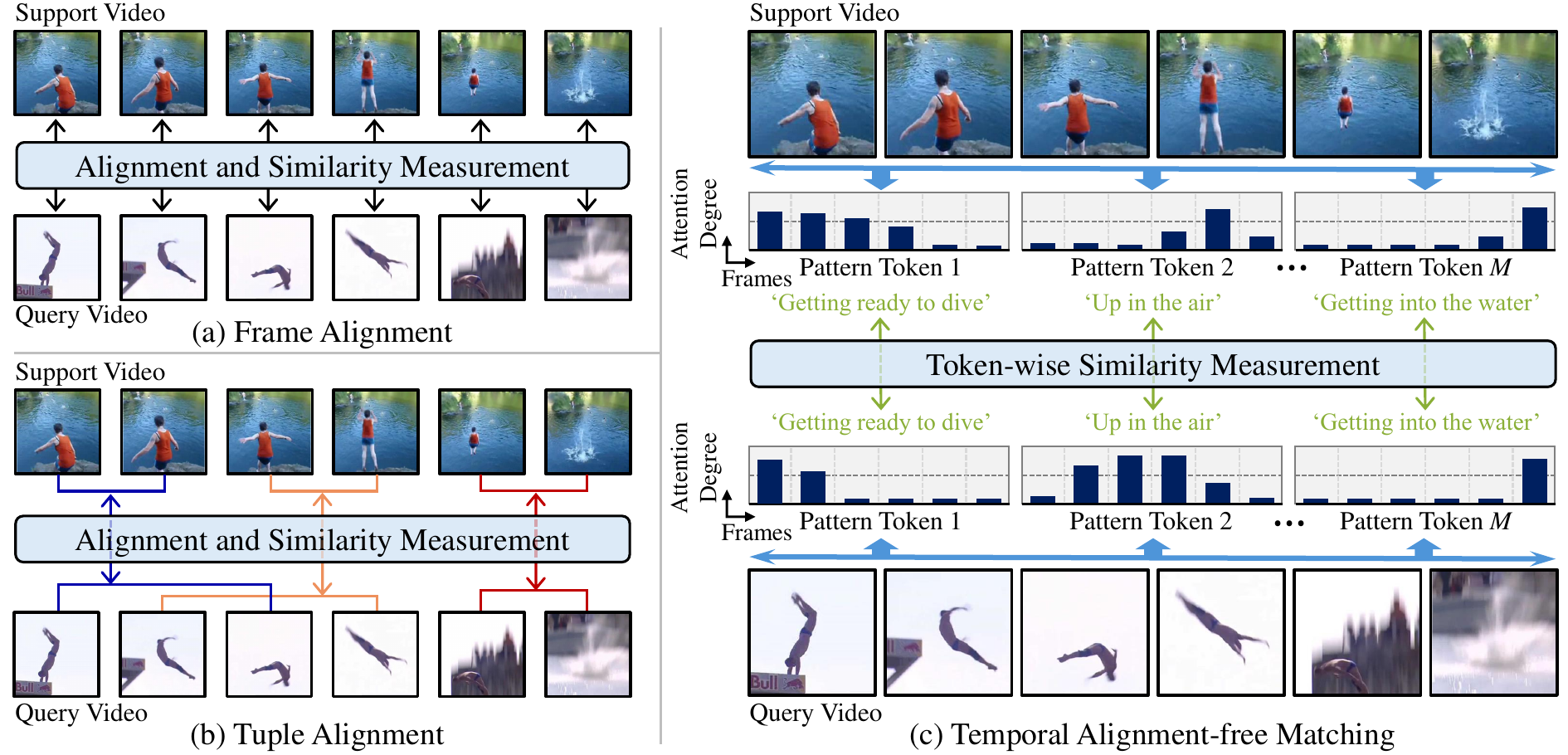} }
    \vspace{-0.6cm}
    \caption{
        Comparison with alignment-based approaches on the ``Diving Cliff" class.
        (a) Frame Alignment: For each frame of the query video, the most corresponding frame in the support video is identified, highlighting the importance of precise frame-level alignment.
        (b) Tuple Alignment: The support and query videos are compared in sub-sequence units to account for variations in action speed. The set of varying tuples is pre-defined.
        (c) Temporal Alignment-free Matching: Both in support and query, the video features are initially integrated using pattern tokens, which encode globally discriminative patterns. 
        Then, it directly compares the corresponding aggregated features of the support and query videos.
        Pattern-based aggregation is both efficient and flexible, as it does not require an alignment process and is unaffected by differences in frame count or action speed.
        Note the text description for each pattern token~(green) is intended to provide an intuitive understanding of what each pattern token represents.
    }
\vspace{-0.3cm}
\label{fig:matchings}
\end{figure*}

As the base units for temporal alignment between videos, the frame and tuple are commonly employed~(illustrated in Fig.~\ref{fig:matchings}~(a) and (b)).
Both frame- and tuple-based alignment approaches measure pairwise similarities between each unit and then derive an optimal alignment path for all units of query video.
This path is regarded as the temporal alignment between support and query videos, the query video is assigned with the class label of the support video with minimum cumulative distance.

Although these two temporal alignment paradigms have shown effectiveness, they suffer from inflexibility and inefficiency.
The inflexibility arises from their reliance on pre-defined units, such as frames or tuples, for action representation.
These methods typically struggle to capture actions if the pre-defined units do not align with the varying action durations.
Moreover, their alignment cost increases quadratically with the number of frames, as shown in Fig.~\ref{fig:computational_time}. 
This issue has been overlooked since typical FSAR benchmarking datasets consist mostly of trimmed, short-term videos.
The quadratic cost arises because their alignment strategies rely on pairwise comparisons between the units of support and query videos.

In this regard, we introduce a novel TEmporal Alignment-free Matching (TEAM) approach that achieves both flexibility and efficiency by eliminating the need for pre-defined temporal units in action representation and brute-force alignment for video comparison.
Specifically, TEAM represents each video with a fixed set of pattern tokens with cross-attention mechanisms to capture diverse clues within the video instance which are useful for distinguishing categories.
Therefore, these pattern tokens flexibly identify discriminative features varying lengths across each video, as illustrated in Fig.~\ref{fig:matchings}~(c).
Furthermore, these pattern tokens are trained to capture globally discriminative clues across the dataset.
This is because all videos leverage the entire set of pattern tokens without an alignment process, rather than relying on specific tokens determined by alignment, as shown in Fig.~\ref{fig:matchings}~(c).
Additionally, this token-wise comparison results in the efficiency of TEAM by eliminating the brute-force alignment process computing pairwise similarities between units for the query video matching.

However, pattern tokens may not be fully discriminative for every episode with novel class compositions. 
To address this, we adapt the class boundaries formed by pattern tokens of support set videos.
Particularly, we identify the patterns that commonly appear across classes and remove them from each support video token.
This process establishes clear class boundaries, improving class discrimination for query videos.

To sum up, our contributions are:
\begin{itemize}
\item We propose Temporal Alignment-free Matching for FSAR equipped with flexibility and efficiency by eliminating the need for pre-defined temporal units in action representation and alignment in video comparison.
\item
We develop an episode adaptation process of encoded pattern tokens.
This establishes clear boundaries within the support set for each class composition by excluding information shared with other classes.
\item Our state-of-the-art performances on various benchmarks for FSAR validate the effectiveness of our TEAM.
\end{itemize}
\section{Related Work}
\begin{figure*}
    \centering
    {\includegraphics[width=1.\textwidth]{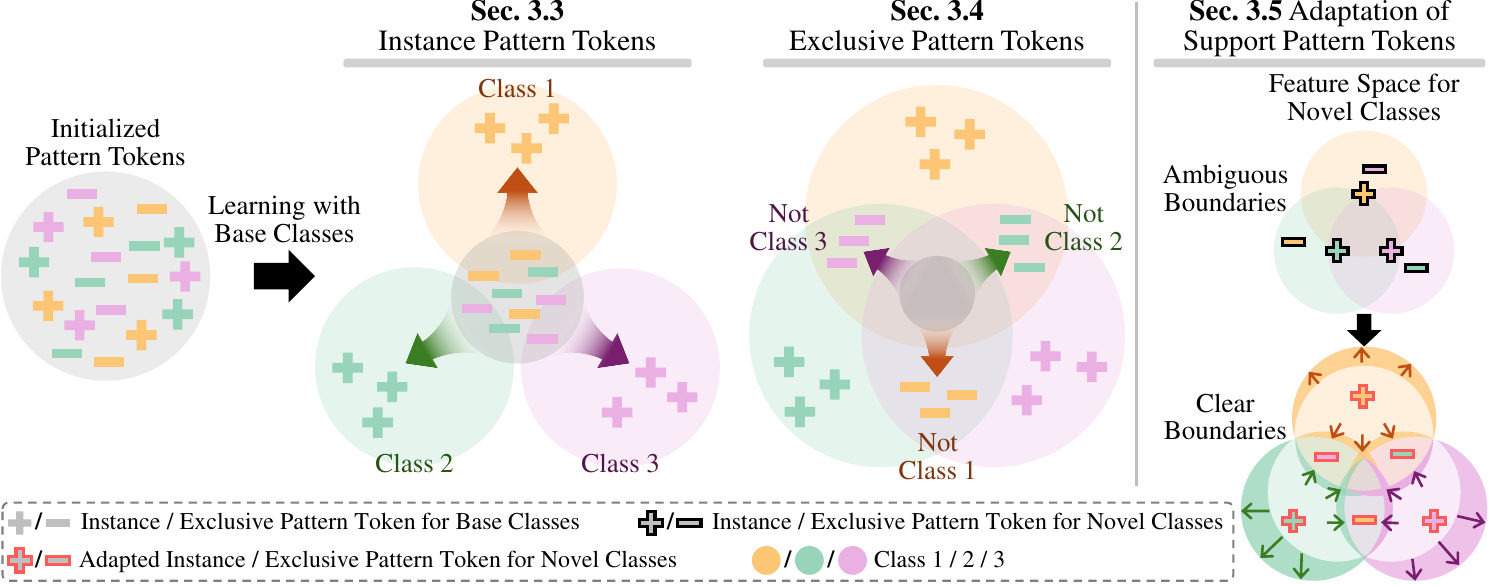} }
    \vspace{-0.6cm}
    \caption{
Illustration of our approach using a single pattern token~($M=1$) in the 3-way 3-shot scenario without query.
This figure illustrates our approach, where instance (+) and exclusive (-) pattern tokens represent integrated video features based on the pattern token.
(Left) Randomly initialized pattern tokens are optimized with classes in two complementary ways.
First, instance pattern tokens~(+) are encouraged to cluster with other instances' tokens of the same class while repelling those of other categories.
On the other hand, exclusive pattern tokens~(-) learn to represent the otherness of each instance by positioning themselves in the embedding space of other classes.
(Right) Although these two types of tokens are discriminative for distinguishing base video categories, they may not fully capture the finer details needed for novel categories.
To address this, we propose an adaptation process of support pattern tokens for novel classes to refine the class decision boundaries.
Note that when multiple pattern tokens are used, these processes run in parallel, with instance and exclusive tokens being compared only within the same pattern token.
    }
\vspace{-0.3cm}
\label{fig:main}
\end{figure*}
\subsection{Few-Shot Action Recognition}
A predominant learning strategy in Few-Shot Action Recognition~(FSAR) is metric learning, which entails inferring the category of a query video based on a pre-defined distance metric~\cite{OTAM, HyRSM, MoLo, TRX, STRM, SloshNet}.
Previous methods focused on measuring the distance between video features through the development of alignment algorithms, either at the frame-level~\cite{OTAM, HyRSM, MoLo} or tuple-level~\cite{TRX, STRM, SloshNet}.
As one of the initial frame alignment methods, OTAM~\cite{OTAM} aligns the temporal order of query and support videos at the frame-level to precisely estimate the distance between potential query-support pairs.
Yet, as it contains a strong assumption that all actions are in a fixed sequence, HyRSM~\cite{HyRSM} proposed a flexible alignment metric alleviating the strictly ordered constraints.
More recently, MoLo~\cite{MoLo} focused on learning long-range temporal perception and motion dynamics through frame-level objectives.
On the other hand, the stream of tuple-level alignment is to gain robustness to varying action lengths and temporal offsets.
Tuple alignment has gained attention for its flexibility in encoding actions of different speeds and temporal offsets~\cite{TRX}.
Additionally, STRM~\cite{STRM} incorporated a spatio-temporal enrichment module to enhance both object and motion representations.
Yet, tuple-based approaches require pre-defined temporal tuples to provide hints for possible forms of actions in videos.
In contrast, our approach is flexible since we do not need pre-defined windows to represent action.

\subsection{Token-based Feature Aggregation}
Feature aggregation generates compact contextual representations, providing both effectiveness and efficiency across various fields~\cite{Perceiver, SlotAttention, ProtoNet, Prost, BLIP2, Flamingo, DETR, TDM, ViT}.
There are various strategies to implement feature aggregation including different pooling methods~\cite{ProtoNet, TDM}.
Among them, our method closely aligns with token-based approaches.
Simply put, the stream of token-based aggregation integrates the input context within the learnable tokens.
For instance, DETR~\cite{DETR} introduces learnable object queries to aggregate spatial features for object detection, while BLIP-2~\cite{BLIP2} and Flamingo~\cite{Flamingo} use learnable tokens to encode image information for multi-modal understanding.
Similarly to prior methods, we use learnable tokens to aggregate video information. 
However, our TEAM differs in that the feature aggregation is implemented in two complementary ways, providing more robust class boundaries.
Additionally, we also propose a token adaptation process to maintain the benefits of capturing discriminative details while encoding instances of novel classes.
\section{Method}

\subsection{Problem Formulation}
In the standard Few-Shot Action Recognition~(FSAR), we are given two sets of data: meta-train set for model training $\mathcal{D}_{base}=\{\left(v_i, y_i\right),y_i\in C_{base}\}$ and meta-test set for model testing $\mathcal{D}_{test}=\{\left(v_i, y_i\right),y_i\in C_{novel}\}$. 
While $v_i$ is a video, $C_{base}$ and $C_{novel}$ denote class sets for meta-training and meta-testing, respectively ($C_{base}\cap C_{novel}=\phi$).
Generally, the training and testing processes of FSAR are composed of episodes where each episode validates the model's adaptability to a novel class set.
Each episode comprises randomly selected $N$ classes and each class consists of $K$ labeled videos and $U$ unlabeled videos, $i.e.$, $N$-way $K$-shot episode.
Whereas a set of labeled videos is called the support set $S={\{\left(v_i, y_i\right)\}}^{N\times K}_{i=1}$, the other set of unlabeled videos are often dubbed as the query set $Q={\{\left(v_i, y_i\right)\}}^{N\times U}_{i=1}$.
In the below sections, we assume $N$-way 1-shot setting where the number of query video $U$ is also set to 1 for better clarity.

\subsection{Overview}
The schematic illustration of our method is shown in \cref{fig:main}.
To eliminate the need for pre-defined units in action representation, we integrate video features using pattern tokens, which capture globally discriminative features across categories.
Specifically, our integration process operates in two complementary ways, generating instance and exclusive pattern tokens.
Instance pattern tokens are designed to capture shared patterns among videos of the same class.
However, these shared patterns may not always be present in every video of the class.
In such cases, relying solely on instance pattern tokens for classification is ineffective.
To address this, exclusive pattern tokens are introduced to represent “otherness”—absent patterns in a given class.
This ensures accurate classification, even when key discriminative patterns are missing from certain videos.
While these pattern tokens are effective for base categories, they may lack the discriminative details for novel classes.
To overcome this, we propose an adaptation process for both instance and exclusive pattern tokens in support videos, reinforcing clear boundaries to distinguish novel classes within each episode.
Notably, pattern tokens are only compared within their type (\textit{i.e.}, token-wise comparison), eliminating the need for an alignment process.

\subsection{Instance Pattern Tokens}
Given uniformly sampled $T$ frames from the support and query videos, we independently extract the features for each frame through the image features extractor, as follows:
\begin{equation}
        \label{eq:feature_extraction}
        {F} = \left[g_\theta({x_1}), g_\theta({x_2}), \cdots, g_\theta({x_T})\right], 
\end{equation}
where $g_\theta$ denotes the image feature extractor $\mathbb{R}^{H\times W \times 3} \rightarrow \mathbb{R}^{D}$ and $x_t$ means $t$-th frame of the video.
$H$ and $W$ are height and width of the frame while $D$ is  size of the features.

Unlike existing works that utilize pre-defined units (\textit{i.e.}, frame or tuple) for representing actions, we integrate video features based on pattern tokens that encode globally discriminative patterns.
Initially, we define a pattern pool that consists of $M$ learnable pattern tokens $P=[P_1, P_2, ..., P_M]$.
Then, we use a cross-attention mechanism to encode instance patterns within $P$.
Formally, instance pattern tokens $P^+ \in \mathbb{R}^{M \times D}$ are computed as:
\begin{equation}
    \begin{split}
        \label{eq:positive_aggregation_2}
        \bar{P}^{+}_m = P_m + \text{CA}(P_m, F, F),
    \end{split}
\end{equation}
\begin{equation}
    \begin{split}
        \label{eq:positive_aggregation_3}
        {P}^{+}_m = \bar{P}^+_m + \text{MLP}(\bar{P}^+_m),
    \end{split}
\end{equation}
where $P_m$ is the $m$-th component of $P$ while $\text{CA}\left(\cdot,\cdot,\cdot\right)$ is the cross-attention receiving \textit{query}, \textit{key}, and \textit{value} as inputs.
And, $\text{MLP}(\cdot)$ indicates a light fully-connected block.

Based on instance pattern tokens of support set $P^{S+}$, the probability for the query $P^{Q+}$ to belong to $n$-th category $p^{+}(y^Q=n | P^{S}, P^{Q})$ is formulated as follows:
\begin{equation}
    \begin{split}
        \label{eq:distance_positive}
        \text{PD}(P^{S}_{n}, P^{Q}) = \sum^M_{m=1} -d(P^{S+}_{n,m}, P^{Q+}_{m}),
    \end{split}
\end{equation}
\begin{equation}
    \begin{split}
        \label{eq:inference_positive}
        p^{+}(y^Q = n | P^{S}, P^{Q}) = \text{softmax}(\text{PD}(P^{S}, P^{Q}); n) ,
    \end{split}
\end{equation}
where $d$ is a distance metric, \textit{i.e.}, the cosine distance, while $P^{S+}_{n,m}$ and $P^{Q+}_{m}$ denote $m$-th instance pattern tokens of support belonging to category $n$ and query, respectively.
In addition, $\text{softmax}(\cdot; n)$ represents the $n$-th component of the softmax function applied to the input.

The key advantage of using pattern tokens is achieving window-free video understanding. 
Unlike previous approaches that represent action with frame- or tuple-based units, pattern tokens are not restricted by specific lengths or speeds.
Furthermore, each pattern token learns globally discriminative features for video classification, since all pattern tokens contribute across the dataset rather than relying on specific tokens determined by alignment.
This design of pattern tokens increases flexibility in action representation and removes the computational cost associated with alignment.

\subsection{Exclusive Pattern Tokens}
In addition to the instance pattern tokens that capture specific patterns in the video, we introduce exclusive pattern tokens, designed to encode the otherness to instance pattern tokens.
In other words, the exclusive tokens are located close to the other classes' instance tokens and differentiated from the instance tokens of the same class.
To emphasize this distinction, we use subtraction rather than addition in Eq.~\ref{eq:positive_aggregation_2} to directly exclude information related to the anchor class, thus the exclusive pattern tokens represent information contrary to instance ones.
Specifically, we define the exclusive-pattern tokens $P^- \in \mathbb{R}^{M \times D}$ as follows:
\begin{equation}
    \begin{split}
        \label{eq:negative_1}
        \bar{P}^{-}_m = P_m - \text{CA}(P_m, F, F),
    \end{split}
\end{equation}
\begin{equation}
    \begin{split}
        \label{eq:negative_2}
        P^{-}_m = \bar{P}^-_m + \text{MLP}(\bar{P}^-_m).
    \end{split}
\end{equation}
Then, we define the probability of the query belonging to the $n$-th category $p^-(y^Q=n | P^{S}, P^{Q})$ with the exclusive and instance pattern tokens, as follows:
\begin{equation}
    \begin{split}
        \label{eq:distance_negative}
        \text{ND}(P^{S}_{n}, P^{Q}) = 
        \underset{\substack{1\leq o \leq N \\ o\neq n}}{\text{min}} 
        \sum^M_{m=1}
        \big( &-d(P^{S-}_{o,m}, P^{Q+}_{m}) 
        \\
        &- d(P^{S+}_{o,m}, P^{Q-}_{m})
        \big),
    \end{split}
\end{equation}
\begin{equation}
    \begin{split}
        \label{eq:inference_negative}
        p^{-}(y^Q=n | P^{S}, P^{Q}) = \text{softmax}(\text{ND}(P^{S}, P^{Q}); n).
    \end{split}
\end{equation}
To accurately estimate the category of the query using the above probability, the exclusive pattern tokens should be similar to the instance pattern tokens of other classes, while being differentiated from those of the same class.
Thus, as learning progresses, the exclusive pattern tokens effectively encode the otherness of the same-class instance pattern tokens, as intended.

\begingroup
\setlength{\tabcolsep}{6pt} 
\renewcommand{\arraystretch}{1} 
\begin{table}[t]
    \centering
    \vspace{-0.3cm}
    \caption{
    The number of pattern tokens.
    }
    \label{tab:number_of_pattern_tokens}
    \vspace{-0.25cm}
    \begin{tabular}{l | c c | c c}
    \hlineB{2.5}
     & \multicolumn{2}{c|}{ResNet} & \multicolumn{2}{c}{ViT} 
    \\
    & 1-shot & 5-shot & 1-shot & 5-shot
    \\
    \hlineB{2.5}
    HMDB51 & 60 & 70 & 50 & 60
    \\
    Kinetics & 60 & 80 & 80 & 80
    \\
    UCF101 & 60 & 80 & 70 & 70
    \\
    SSv2-Small& - & - & 50 & 80
    \\
    \hlineB{2.5}
    \end{tabular}
    \vspace{-0.1cm}
\end{table}
\endgroup
\subsection{Adaptation of Support Pattern Tokens}
Although the pattern tokens learn globally discriminative patterns across base categories in the training phase, they might not be as effective for novel classes.
Therefore, we propose an adaptation process for the support set to achieve a clear separation between categories within an episode. 

Specifically, we remove common information shared with other classes from instance pattern tokens and allocate the discriminative information of other classes into exclusive pattern tokens.
To regulate the degree of adaptation for each episode, we use the cosine similarity between pattern tokens of the anchor class and other classes, representing the level of semantic entanglement. 
The adaptive instance pattern tokens $\hat{P}^{S+} \in \mathbb{R}^{M \times D}$ are defined as follows:
\begin{equation}
    \begin{split}
        \label{eq:episode_adaptive_positive_1}
        E^+_{n,o,m} = P^{S+}_{n,m} \cdot P^{S+}_{o,m},
    \end{split}
\end{equation}
\begin{equation}
    \begin{split}
        \label{eq:episode_adaptive_positive_2}
        \tilde{P}^{S+}_{n,o,m} = P_m + (1 + E^+_{n,o,m})&\text{CA}(P_m, F^S_n, F^S_n)
        \\
        - E^+_{n,o,m}&\text{CA}(P_m, F^S_o, F^S_o),
    \end{split}
\end{equation}
\vspace{-0.2cm}
\begin{equation}
    \begin{split}
        \label{eq:episode_adaptive_positive_3}
        \!\!\!
        \hat{P}^{S+}_{n,m} = \frac{1}{N-1}\sum^N_{o=1}
        \mathbbm{1}_{\left[n \neq o\right]}\big( \tilde{P}^{S+}_{n,o,m} + \text{MLP}(\tilde{P}^{S+}_{n,o,m}) \big),
    \end{split}
\end{equation}
where $\left( \cdot \right)$ means the cosine similarity measure between two vectors, and $P^{S+}_{n,m}$ is $m$-th instance pattern token of $n$-th class in the support set, while $\mathbbm{1}_{\left[n \neq o\right]}$ is an indicator function that outputs 1 if $n$ is different with $o$ and 0 otherwise.

Similarly, the adaptive pattern tokens for exclusive ones $\hat{P}^{S-} \in \mathbb{R}^{M \times D}$ are formulated as follows:
\begin{equation}
    \begin{split}
        \label{eq:episode_adaptive_negative_2}
        \tilde{P}^{S-}_{n,o,m} = P_m - (1 + E^-_{n,o,m}) &\text{CA}(P_m, F^S_n, F^S_n)
        \\
        + E^-_{n,o,m}&\text{CA}(P_m, F^S_o, F^S_o),
    \end{split}
\end{equation}
\vspace{-0.2cm}
\begin{equation}
    \begin{split}
        \label{eq:episode_adaptive_negative_3}
        \!\!\!
        \hat{P}^{S-}_{n,m} = \frac{1}{N-1}\sum^N_{o=1}
        \mathbbm{1}_{\left[n \neq o\right]}\big( \tilde{P}^{S-}_{n,o,m} + \text{MLP}(\tilde{P}^{S-}_{n,o,m}) \big).
    \end{split}
\end{equation}

Eq.~\ref{eq:episode_adaptive_positive_2} extends Eq.~\ref{eq:positive_aggregation_2} by introducing a process to remove shared information between classes while generating instance tokens.
This process suppresses information shared between the anchor class $n$ and the other class $o$.
It achieves this by subtracting $\text{CA}(P_m, F^S_o, F^S_o)$ from $\text{CA}(P_m, F^S_n, F^S_n)$, thereby retaining only the discriminative information for the anchor class $n$.
The extent of suppression is proportional to the degree of information overlap, controlled by $E^+_{n,o,m}$.
This ensures that the adaptive instance pattern tokens focus on the discriminative features of novel classes.
Similarly, Eq.~\ref{eq:episode_adaptive_negative_2} follows the same principle but instead emphasizes the discriminative features of other classes.

\subsection{Training Objective and Inference}
Adjusting the probabilities defined in Eq.~\ref{eq:inference_positive} and \ref{eq:inference_negative} with adaptive pattern tokens for the support set, we define two objective functions for model training as follows:
\begin{equation}
    \begin{split}
        \label{eq:loss_positive}
        \mathcal{L}^+=-\mathbbm{1}_{\left[y^Q=n\right]} \text{log}
        \big( p^{+}(y^Q=n | \hat{P}^{S}, P^{Q})
        \big),
    \end{split}
\end{equation}
\begin{equation}
    \begin{split}
        \label{eq:loss_negative}
        \mathcal{L}^-=-\mathbbm{1}_{\left[y^Q=n\right]} \text{log}
        \big( 
        p^{-}(y^Q=n | \hat{P}^{S}, P^{Q})
        \big).
    \end{split}
\end{equation}
Then, the final loss function $\mathcal{L}$ is defined as follows:
\begin{equation}
    \begin{split}
        \label{eq:loss_final}
        \mathcal{L}=\mathcal{L}^+ + \mathcal{L}^-.
    \end{split}
\end{equation}
In short, instance pattern tokens are optimized to resemble those of the same class and exclusive tokens of other classes. 
At the same time, they remain distinct from instance tokens of other classes and exclusive tokens of the same class.

As a result, the probability of the query to belong to $n$-th class $p(y^Q=n)$ is defined as follows:
\begin{equation}
    \begin{split}
        \label{eq:inference_final}
        \!\!\!\!\!
        p(y^Q \!=\!n) \!=\! \text{softmax}\big(
        \text{PD}(\hat{P}^{S} \!\!, P^{Q}) + \text{ND}(\hat{P}^{S} \!\!, P^{Q}); n\big).
    \end{split}
\end{equation}
\section{Experiments}\label{sec:experiments}
\paragraph{Datasets.}
We conduct experiments on four benchmark datasets for Few-Shot Action Recognition (FSAR): HMDB51~\cite{HMDB51}, Kinetics~\cite{Kinetics}, UCF101~\cite{UCF101}, and SSv2-Small~\cite{SSv2}.
For HMDB51, we adopt the settings from ARN~\cite{ARN}, using 31, 10, and 10 classes for training, validation, and testing, respectively.
Kinetics is adapted to the few-shot setting following previous works~\cite{OTAM, CMN_Extension}, with 64, 12, and 24 categories assigned to training, validation, and testing.
For UCF101, we follow the ARN~\cite{ARN} settings, using 70, 10, and 21 classes for training, validation, and testing.
Lastly, for SSv2-Small, we use the split settings from CMN~\cite{CMN}, with 64, 12, and 24 categories for training, validation, and testing.

\begingroup
\setlength{\tabcolsep}{10pt} 
\renewcommand{\arraystretch}{1.0} 
\begin{table*}[t]
    \centering
    \caption{
Performance comparison on FSAR benchmark datasets under the 5-way setting. 
The best results are \textbf{bolded}, while the second-best are \underline{underlined}. $
\dagger$ indicates that evaluation was conducted in a transductive setting, utilizing relations between queries.
$\ast$ denotes reproduced performances, as results with ViT were unavailable.
    }\label{tab:main}
    \vspace{-0.2cm}
    \begin{tabular}{l | c | c | c c | c c | c c}
        \hlineB{2.5}
        \multirow{2}{*}{Model} &
        \multirow{2}{*}{Reference} &
        \multirow{2}{*}{Backbone} & 
        \multicolumn{2}{c|}{HMDB51} &
        \multicolumn{2}{c|}{Kinetics} &
        \multicolumn{2}{c}{UCF101}
        \\
        & & & 1-shot & 5-shot & 1-shot & 5-shot & 1-shot & 5-shot
        \\
        \hlineB{2.5}
        ProtoNet~\cite{ProtoNet} & NeurIPS 2017 & \multirow{16}{*}{ResNet} & 54.2 & 68.4 & 64.5 & 77.9 & 74.0 & 89.6
        \\
        CMN~\cite{CMN} & ECCV 2018 & & - & - & 57.3 & 76.0 & - & -
        \\
        OTAM~\cite{OTAM} & CVPR 2020 & & 54.5 & 68.0 & 72.2 & 84.2 & 79.9 & 88.9
        \\
        ARN~\cite{ARN} & ECCV 2020 & & 45.5 & 60.6 & 63.7 & 82.4 & 66.3 & 83.1
        \\
        TRX~\cite{TRX} & CVPR 2021 & & 53.1 & 75.6 & 63.6 & 85.9 & 78.2 & 96.1
        \\
        TA2N~\cite{TA2N} & AAAI 2022 & & 59.7 & 73.9 & 72.8 & 85.8 & 81.9 & 95.1
        \\
        STRM~\cite{STRM} & CVPR 2022 & & 52.3 & 77.3 & 62.9 & 86.7 & 80.5 & \underline{96.9}
        \\
        HyRSM~\cite{HyRSM} & CVPR 2022 & & 60.3 & 76.0 & 73.7 & 86.1 & 83.9 & 94.7
        \\
        MTFAN~\cite{MTFAN} & CVPR 2022 & & 59.0 & 74.6 & 74.6 & \underline{87.4} & 84.8 & 95.1
        \\
        HCL~\cite{HCL} & ECCV 2022 & & 59.1 & 76.3 & 73.7 & 85.8 & 82.5 & 93.9
        \\
        \textit{Nguyen et al.}~\cite{Nguyen} & ECCV 2022 & & 59.6 & 76.9 & 74.3 & \underline{87.4} & 84.9 & 95.9
        \\
        \textit{Huang et al.}~\cite{Huang} & ECCV 2022 & & 60.1 & 77.0 & 73.3 & 86.4 & 71.4 & 91.0
        \\
        SloshNet~\cite{SloshNet} & AAAI 2023 & & - & \underline{77.5} & - & 87.0 & - & \textbf{97.1}
        \\
        MoLo~\cite{MoLo} & CVPR 2023 & & 60.8 & 77.4 & 74.0 & 85.6 & \underline{86.0} & 95.5
        \\
        $\text{GgHM}^\dagger$~\cite{GgHM} & ICCV 2023 & & \underline{61.2} & 76.9 & \underline{74.9} & \underline{87.4} & 85.2 & 96.3
        \\
        \rowcolor{gray!30} TEAM & CVPR 2025 & & \textbf{62.8} & \textbf{78.4} & \textbf{75.1} & \textbf{88.2} & \textbf{87.2} & 96.2
        \\
        \hlineB{2.0}
        OTAM$^\ast$~\cite{OTAM} & CVPR 2020 & \multirow{6}{*}{ViT} & 69.3 & 83.1 & \underline{82.2} & \underline{91.6} & 92.9 & 97.5
        \\
        TRX$^\ast$~\cite{TRX} & CVPR 2021 & & 61.3 & 82.2 & 76.8 & 90.4 & 85.8 & \underline{97.8}
        \\
        STRM$^\ast$~\cite{STRM} & CVPR 2022 & & 56.7 & \underline{83.2} & 75.8 & 90.8 & 90.8 & 97.6
        \\
        MoLo$^\ast$~\cite{MoLo} & CVPR 2023 & & \underline{69.9} & \underline{83.2} & 82.0 & 91.2 & \underline{93.6} & \underline{97.8}
        \\
        TATs~\cite{TATs} & ECCV 2024 & & 60.0 & 77.0 & 81.9 & 91.1 & 92.0 & 95.5
        \\
        \rowcolor{gray!30} TEAM & CVPR 2025 & & \textbf{70.9} & \textbf{85.5} & \textbf{83.3} & \textbf{92.9} & \textbf{94.5} & \textbf{98.8}
        \\
        \hlineB{2.5}
    \end{tabular}
    \vspace{-0.4cm}
\end{table*}
\endgroup

\begingroup
\setlength{\tabcolsep}{8.5pt} 
\renewcommand{\arraystretch}{1.0} 
\begin{table}[t]
    \centering
    \caption{
Performance comparison in SSv2-Small dataset using ViT.
TATs w/o PT refers to TATs without the additional off-the-shelf Point Tracker.
Symbols match those in Tab.~\ref{tab:main}
    }
    \label{tab:temporally_difficult_action}
    \vspace{-0.2cm}
    \begin{tabular}{l | c | c c }
        \hlineB{2.5}
        \multirow{2}{*}{Model} & \multirow{2}{*}{Reference} & \multicolumn{2}{c}{SSv2-Small}
        \\
        & & 1-shot & 5-shot
        \\
        \hlineB{2.5}
        OTAM$^\ast$~\cite{OTAM} & CVPR 2020 & 44.4 & 58.2
        \\
        TRX$^\ast$~\cite{TRX} & CVPR 2021 & 42.0 & 59.1 
        \\
        STRM$^\ast$~\cite{STRM} & CVPR 2022 & 42.5 & 60.4
        \\
        MoLo$^\ast$~\cite{MoLo} & CVPR 2023 & 45.1 & 58.8
        \\
        TATs~\cite{TATs} & ECCV 2024 & \textbf{47.9} & \textbf{64.4}
        \\
        TATs w/o PT~\cite{TATs} & ECCV 2024 & 44.6 & 62.5
        \\
        \rowcolor{gray!30} TEAM & CVPR 2025 & \underline{47.2} & \underline{63.1}
        \\
        \hlineB{2.5}
    \end{tabular}
    \vspace{-0.2cm}
\end{table}
\endgroup 
\paragraph{Implementation Details.}
\label{sec:implementation_details}
For a fair comparison with existing works, we follow existing protocols~\cite{CMN, OTAM, MoLo, TATs}.
We use ResNet-50~\cite{resnet} and ViT-B~\cite{ViT} as a backbone network, both pre-trained on ImageNet~\cite{imagenet}.
The backbone processes each video by taking 8 uniformly sampled frames at a resolution of $224\times224$ as input (\textit{i.e.}, $T=8$).
As in prior studies, data augmentation techniques such as random cropping and color jittering are applied~\cite{HyRSM, MoLo, GgHM}.
The number of learnable pattern tokens is listed in Tab.~\ref{tab:number_of_pattern_tokens}.
For the optimization, we adopt SGD to train our model for 10,000 iterations on all datasets.
In the many-shot scenarios, we utilize the prototype concept~\cite{ProtoNet}, following MoLo~\cite{MoLo}.
We measure the performance of our model as the average results over 10,000 randomly sampled episodes.
All experiments are conducted on RTX A6000 GPUs with Pytorch Cuda amp.

\subsection{Comparison with state-of-the-art methods}
In Tab.~\ref{tab:main}, we compare our proposed method, TEAM, with previous FSAR approaches under the 5-way 1-shot and 5-shot scenarios, which are standard evaluation protocols.
Using ResNet~\cite{resnet}, a conventional backbone network, TEAM achieves state-of-the-art performance across various datasets and shot numbers, except for UCF101~(5-shot).
With the ViT backbone~\cite{ViT}, TEAM surpasses all other methods in every case.
We attribute TEAM’s effectiveness to its use of globally shared pattern tokens, which inherently capture discriminative patterns across diverse scenarios.
This approach enables both efficient and reliable adaptation to new episodes by leveraging pattern tokens to compare videos through multiple criteria.
Additionally, TEAM is not only effective but also significantly more efficient than frame- and tuple-level alignment methods, as shown in Fig.~\ref{fig:computational_time}.
These results demonstrate that TEAM is a robust alternative to existing alignment strategies in general FSAR settings.

\begingroup
\setlength{\tabcolsep}{7.8pt} 
\renewcommand{\arraystretch}{1.0} 
\begin{table}[t]
    \centering
    \caption{
Performance comparison across cross-domain scenarios. 
All models are trained on the 5-way 5-shot scenario using the Kinetics dataset with the ViT backbone, and evaluated on the same task across various datasets.
Symbols match those in Tab.~\ref{tab:main}.
    }
    \label{tab:cross_domain}
    \vspace{-0.2cm}
    \begin{tabular}{l | c | c | c }
        \hlineB{2.5}
        Model & Reference & HMDB51 & UCF101
        \\
        \hlineB{2.5}
        OTAM$^\ast$~\cite{OTAM} & CVPR 2020 & 77.2 & 96.5
        \\
        TRX$^\ast$~\cite{TRX} & CVPR 2021 & 78.1 & 97.9
        \\
        STRM$^\ast$~\cite{STRM} & CVPR 2022 & \underline{78.5} & \underline{98.0}
        \\
        MoLo$^\ast$~\cite{MoLo} & CVPR 2023 & 77.7 & 97.4
        \\
        \rowcolor{gray!30} TEAM & CVPR 2025 & \textbf{81.1} & \textbf{98.3}
        \\
        \hlineB{2.5}
    \end{tabular}
    \vspace{-0.4cm}
\end{table}
\endgroup 
We further conduct experiments on SSv2-Small, a benchmark with high temporal complexity and subtle motion differences, to validate TEAM's effectiveness in handling complex temporal dynamics.
As shown in Tab.~\ref{tab:temporally_difficult_action}, TEAM surpasses other methods, except for TATs which use an additional off-the-shelf point tracker~(PT).
Without PT (TATs w/o PT) which generates tailored object trajectories for temporal actions, TEAM also outperforms TATs even in SSv2-Small.
These results suggest that the benefits of TEAM remain robust in the context of temporally challenging actions.

Cross-domain evaluation is widely used to assess a model’s generalization capability~\cite{Kinetics, cross_domain_1, cross_domain_2, cross_domain_3}.
It typically leverages knowledge learned from coarse datasets to recognize actions in more specific and nuanced domains, highlighting the model's adaptability.
Following this protocol, we compare TEAM with other FSAR methods in Tab.~\ref{tab:cross_domain}.
We train all models on 5-way 5-shot tasks using the Kinetics dataset with the ViT backbone and evaluate them on the same setting across different target datasets.
As shown in Tab.~\ref{tab:cross_domain}, TEAM consistently outperforms prior methods, indicating strong generalization beyond the training domain.
This demonstrates TEAM’s robustness to temporal variations with pattern tokens capturing globally distinct patterns.

\subsection{Ablation Study}
\label{sec:ablation_study}
Unless specifically stated, experiments are conducted in the 5-way 1-shot scenario using the ResNet-50.


\begingroup
\setlength{\tabcolsep}{7.9pt} 
\renewcommand{\arraystretch}{1.0} 
\begin{table}[t]
    \centering
    \caption{
Ablation study for each component. 
$P^+$, $P^-$, and $\hat{P}$ denote the instance, exclusive, adaptive pattern tokens, respectively, while $E$ means entanglement in Eq.~\ref{eq:episode_adaptive_positive_1} and \ref{eq:episode_adaptive_negative_2}.
H, K, and U also refer to HMDB51, Kinetics, and UCF101, respectively.
    }
    \label{tab:ablation}
    \vspace{-0.15cm}
    \begin{tabular}{c | c c c c | c | c | c}
        \hlineB{2.5}
        &
        $P^+$ &
        $P^-$ &
        $\hat{P}$ &
        $E$ &
        H &
        K &
        U
        \\
        \hlineB{2.5}
        (a) & \checkmark & & & & 61.8 & 74.6 & 86.7
        \\
        (b) & \checkmark & \checkmark & & & 62.5 & 75.0 & 86.8
        \\
        (c) & \checkmark & \checkmark & \checkmark & & 62.2& 74.8 & 87.1
        \\
        (d) & \checkmark & \checkmark & \checkmark & \checkmark & \textbf{62.8} & \textbf{75.1} & \textbf{87.2}
        \\
        \hlineB{2.5}
    \end{tabular}
    \vspace{-0.3cm}
\end{table}
\endgroup

\vspace{-0.1cm}
\paragraph{Component Ablation.}
Tab.~\ref{tab:ablation} presents the performance when our component is applied sequentially.
(a) shows that the use of instance pattern tokens solely achieves decent performances over the methods that rely on pre-defined temporal units and brute-force alignment processes.
This validates the advantage of our core intuition that temporal alignment-free matching is beneficial for its flexibility in video understanding.
Succeedingly, in (b), the exclusive tokens are shown to yield additional improvements, indicating that the different strategy in learning pattern tokens complements the original approach.
Conversely, we observe slight performance drops in (c) where we implement the adaptation process without the management in the adaptation magnitude.
This implies the importance of managing the degree of adaptation with the consideration of given class compositions.
The performance gains in (d) validate our claim.

\begin{figure}
    \centering
    {\includegraphics[width=\linewidth]{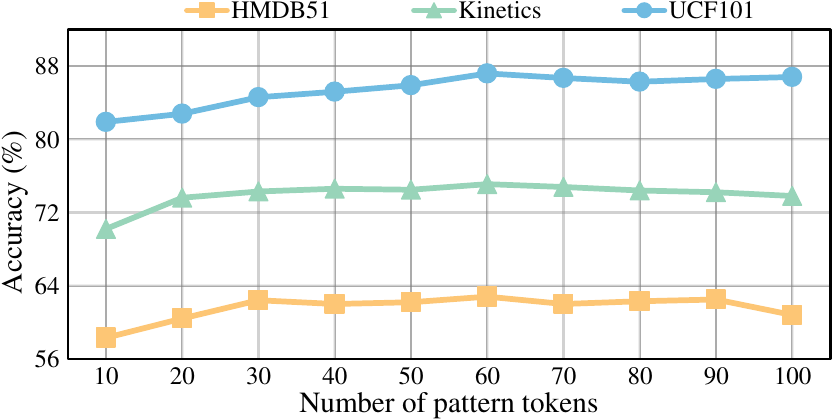} }
    \vspace{-0.6cm}
    \caption{
        Ablation study for the number of pattern tokens.
    }
    \vspace{-0.3cm}
\label{fig:number_of_pattern_tokens}
\end{figure}

\vspace{-0.1cm}
\paragraph{Number of pattern tokens.}
In Fig.~\ref{fig:number_of_pattern_tokens}, we present an ablation study on the effect of the number of pattern tokens in TEAM.
The results show that TEAM remains robust across a wide range of token counts, except when the value deviates significantly from the optimal~(60 for all datasets).
These consistent trends across all datasets suggest that our method is not sensitive to the number of pattern tokens, demonstrating its strong robustness.

\begin{figure}
    \centering
    {\includegraphics[width=\linewidth]{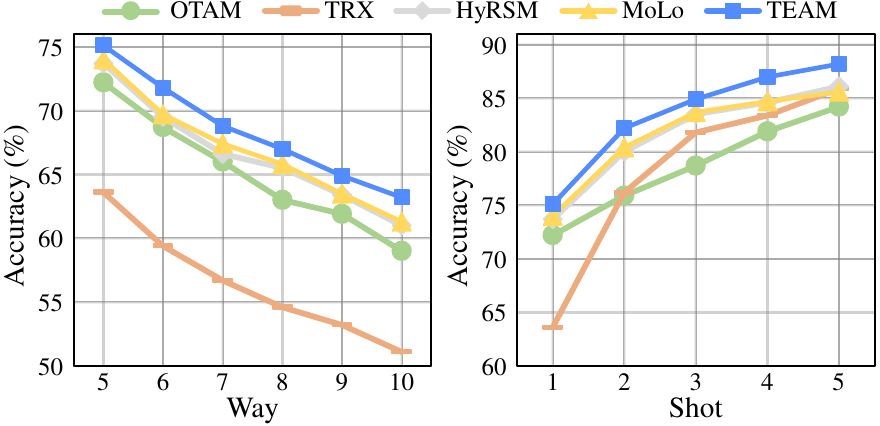} }
    \vspace{-0.6cm}
    \caption{
        N-way 1-shot and 5-way K-shot results.
    }
    \vspace{-0.3cm}
\label{fig:ablation_wayshot}
\end{figure}
\paragraph{Varying N and K for N-way K-shot.}
Beyond standard FSAR settings, we further validate TEAM’s effectiveness under varying N-way K-shot configurations on Kinetics, following prior work~\cite{MoLo}.
Specifically, we vary the number of classes $N$ from 5 to 10, and the number of video instances per class $K$ from 1 to 5.
As shown in Fig.~\ref{fig:ablation_wayshot}, TEAM consistently outperforms prior approaches across all configurations.
These results confirm that TEAM is robust to different FSAR settings, with consistent benefits across diverse scenarios.

\begin{figure*}[t]
    \centering
    {\includegraphics[width=1.0\textwidth]{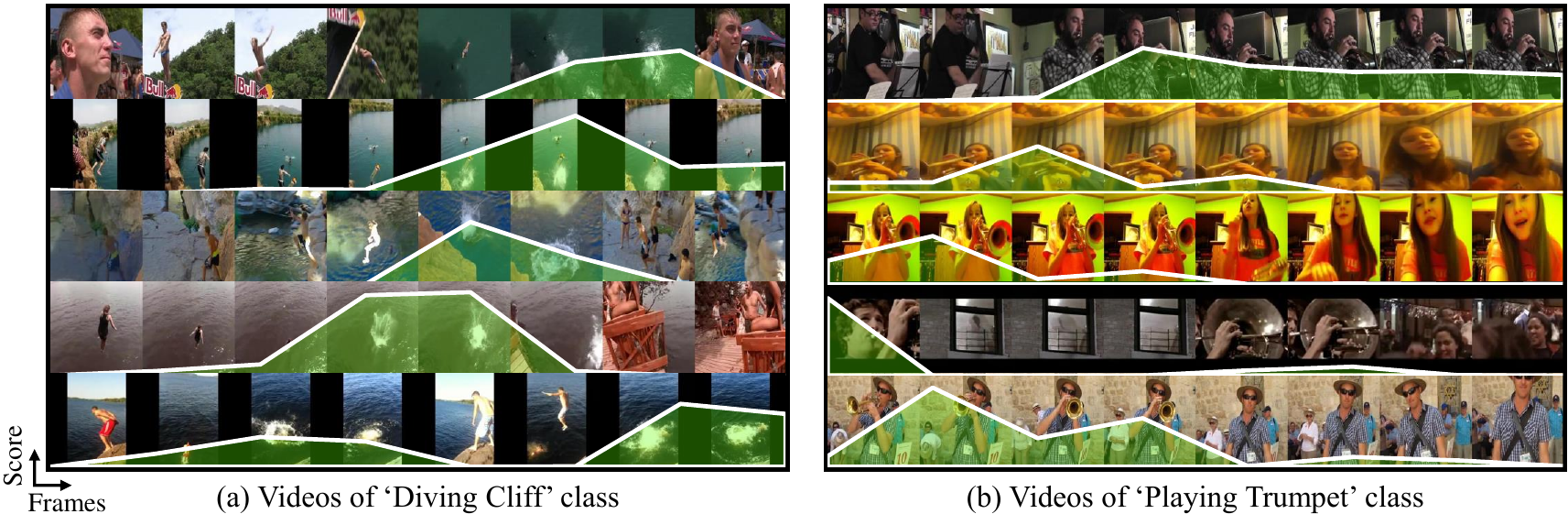} }
    \vspace{-0.6cm}
    \caption{
    Visualization of cross-attention weights during instance pattern token aggregation.
    The same pattern tokens are visualized for videos in each class.
    Green regions in each frame metaphorically represent the percentage of the frame involvement in the instance pattern token.
    (a) For `Diving Cliff', the visualized token primarily responds to the moment when water splashes when people dive.
(b) In `Playing Trumpet', the token focuses on the scenes where people hold a trumpet to their mouths.
    }
\label{fig:qualitative_results}
\end{figure*}
\subsection{Analysis}
\paragraph{What do pattern tokens encode?}
Pattern tokens are leveraged to incorporate video features and have been verified to be superior. 
To assess the video understanding capability of pattern tokens, we examine whether pattern tokens focus on the same context for the same category. 
In Fig.~\ref{fig:qualitative_results}, we visualize the attention weights of a pattern token that effectively represents specific classes. 
Pattern tokens are considered effective in representing a class when the output pattern tokens for instances of the same class exhibit the highest similarity among all pattern tokens. 
As observed, pattern tokens consistently focus on the context even when the same action occurs in different frames, at varying speeds, and over different lengths. 
This implies that pattern tokens address the limitations of frame- and tuple-level alignment approaches, where actions can only be compared within pre-defined units.

\vspace{-0.3cm}
\paragraph{Necessity for multiple pattern tokens.}
\begin{figure}
    \centering
    \vspace{-0.3cm}
    {\includegraphics[width=\linewidth]{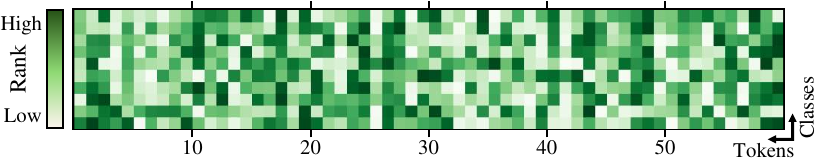} }
    \vspace{-0.75cm}
    \caption{
Heatmap visualizing the class-wise discriminative power of each instance pattern token for novel classes within HMDB51. 
Differences in discrimination ranks across classes indicate that each class has unique discriminative patterns.
    }
\label{fig:heatmap}
\end{figure}
We adopt multiple pattern tokens to capture diverse discriminative patterns across base categories, using these patterns as a foundation for distinguishing novel classes. 
To verify that the learned patterns effectively differentiate novel classes, we compute the discriminative power of each instance pattern token within each category and rank the tokens accordingly.
Discriminative power is defined as `(intra similarity - inter similarity)'.
Intra similarity measures the average similarity between instance-specific pattern tokens from all videos within the same category and their prototype (average), while inter similarity is the maximum similarity between the prototype of one class and the prototypes of other classes. 
The resulting heatmap of discriminative power is shown in Fig.~\ref{fig:heatmap}, where distinct heatmap arrangements between classes can be observed.
The variation in discriminative rank indicates that useful patterns differ by category, supporting the need for multiple learnable pattern tokens to capture a variety of discriminative patterns, forming a foundation for classifying novel categories.

\paragraph{Computational time for matching process.}
\begin{figure}
    \centering
    \vspace{-0.4cm}
    {\includegraphics[width=\linewidth]{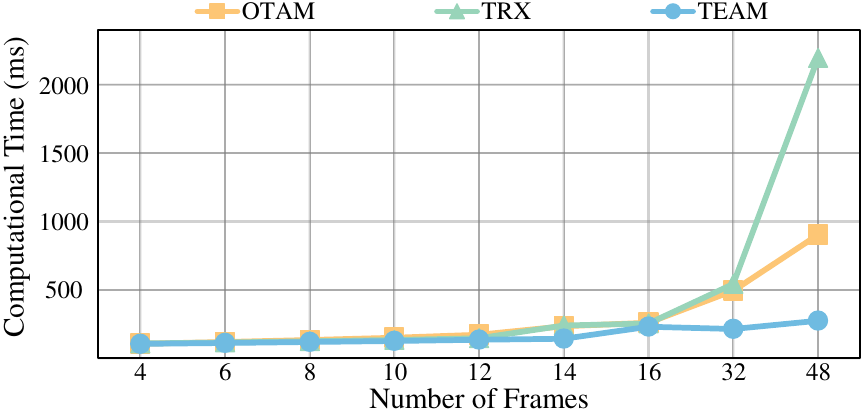} }
    \vspace{-0.6cm}
    \caption{
Comparison of computational time~(ms) for video matching with alignment-based strategies across varying frame counts.
OTAM and TRX are used as representatives of the frame- and tuple-level alignment approaches, respectively.
Times reflect only the video matching process and exclude feature extraction by the backbone network, which is consistent across all methods.
    }
    \vspace{-0.2cm}
\label{fig:computational_time}
\end{figure}
In this study, we compare the computational overhead with alignment-based approaches. 
Computational costs for varying numbers of frames are shown in Fig.~\ref{fig:computational_time}. 
As discussed in Sec.~\ref{sec:intro}, alignment-based strategies require quadratic computation time due to pairwise comparison between temporal units, whereas TEAM with token-wise comparison has linear time complexity—especially evident when the number of frames exceeds 32.
Although this issue is less critical for FSAR datasets, which consist of trimmed, short-term videos, it becomes significant in real-world scenarios with untrimmed, long-term videos. 
In this regard, we argue that our proposed temporal alignment-free matching offers a compelling alternative to alignment-based approaches.

\paragraph{FLOPs and parameter.}
We analyze FLOPs and parameter counts of TEAM and other methods, as shown in Tab.~\ref{tab:flops_parameter}.
While TEAM achieves remarkable performance, it has the fewest parameters except for OTAM, which relies solely on backbone parameters.
The higher parameter counts in other methods stem from their complex designs, including temporal transformers.
These results highlight the effectiveness of pattern tokens for FSAR, providing a lightweight alternative to heavier models.
Moreover, although TEAM has higher FLOPs than OTAM and TRX, its efficient parallel operations ensure low computational time, as shown in Fig.~\ref{fig:computational_time}.
\begingroup
\setlength{\tabcolsep}{4.25pt} 
\renewcommand{\arraystretch}{1.0} 
\begin{table}[t]
    \centering
    \vspace{-0.3cm}
    \caption{
    Comparison of FLOPs and parameter counts, which are measured in G~(giga operations) and M~(million), respectively.
    }
    \label{tab:flops_parameter}
    \vspace{-0.2cm}
    \begin{tabular}{l | c c | c c}
    \hlineB{2.5}
    \multirow{2}{*}{Model} & \multicolumn{2}{c|}{ResNet} & \multicolumn{2}{c}{ViT} 
    \\
    & FLOPs & Parameter & FLOPs & Parameter
    \\
    \hlineB{2.5}
    OTAM~\cite{OTAM} & 198 & 23.5 & 809 & 85.8
    \\
    TRX~\cite{TRX} & 205 & 47.1 & 810 & 88.8 
    \\
    STRM~\cite{STRM} & 201 & 45.5 & 810 & 88.8
    \\
    SloshNet~\cite{SloshNet} & 594 & 168.9 & - & -
    \\
    MoLo~\cite{MoLo} & 237 & 89.5 & 830 & 99.1 
    \\
    TATs~\cite{TATs} & - & - & 1178 & 97.7
    \\
    \rowcolor{gray!30} TEAM & 224 & 32.0 & 813 & 87.0
    \\
    \hlineB{2.5}
    \end{tabular}
    \vspace{-0.2cm}
\end{table}
\endgroup

\section{Conclusion}
In this paper, we proposed a novel TEmporal Alignment-free Matching~(TEAM) for Few-Shot Action Recognition, that achieves both flexibility and efficiency by eliminating the need for pre-defined temporal units in action representation and brute-force alignment for the video comparison.
Concretely, TEAM integrates each video with a fixed set of pattern tokens that encapsulate globally discriminative patterns within the video instance, regardless of action length and speed.
Moreover, TEAM has inherent efficiency by computing token-wise similarity for matching video.
Upon these pattern tokens, we further proposed an adaptation process for establishing clear boundaries between classes by removing common information.
We validated the effectiveness of TEAM with extensive experiments.

\paragraph{Acknowledgements}
This work was supported in part by MSIT/IITP (No. 2022-0-00680, 2020-0-01821, RS-2019-II190421, RS-2024-00459618, RS-2024-00360227, RS-2024-00437102, RS-2024-00437633), and MSIT/NRF (No. RS-2024-00357729).

{
    \small
    \bibliographystyle{ieeenat_fullname}
    \bibliography{main}
}

\end{document}